# Train Your Data Processor: Distribution-Aware and Error-Compensation Coordinate Decoding for Human Pose Estimation

Feiyu Yang[1,2*], Zhan Song[1], Zhenzhong Xiao[2], Yu Chen[2], Zhe Pan[2], Min Zhang[2],
Min Xue[2], Yaoyang Mo[2], Yao Zhang[2], Guoxiong Guan[2], Beibei Qian[2]

[1] Shenzhen Institutes of Advanced Technology, Chinese Academy of Sciences, Shenzhen 518055, China
[2] Orbbec Inc., Shenzhen 518061, China
hansen@orbbec.com

## Abstract

Recently, the leading performance of human pose estimation is dominated by heatmap based methods. While being a fundamental component of heatmap processing, heatmap decoding (*i.e.* transforming heatmaps to coordinates) receives only limited investigations, to our best knowledge. This work fills the gap by studying the heatmap decoding processing with a particular focus on the errors introduced throughout the prediction process. We found that the errors of heatmap based methods are surprisingly significant, which nevertheless was universally ignored before. In view of the discovered importance, we further reveal the intrinsic limitations of the previous widely used heatmap decoding methods and thereout propose a ***D**istribution-**A**ware and **E**rror-Compensation Coordinate Decoding* (DAEC). Serving as a model-agnostic plug-in, DAEC learns its decoding strategy from training data and remarkably improves the performance of a variety of state-of-the-art human pose estimation models with negligible extra computation. Specifically, equipped with DAEC, the SimpleBaseline-ResNet152-256×192 and HRNet-W48-256×192 are significantly improved by ***2.6 AP*** and ***2.9 AP*** achieving 72.6 AP and 75.7 AP on COCO, respectively. Moreover, the HRNet-W32-256×256 and ResNet-152-256×256 frameworks enjoy even more dramatic promotions of ***8.4%*** and ***7.8%*** on MPII with PCKh$^{0.1}$ metric. Extensive experiments performed on these two common benchmarks demonstrates that DAEC exceeds its competitors by considerable margins, backing up the rationality and generality of our novel heatmap decoding idea. The project is available at *https://github.com/fyang235/DAEC*.

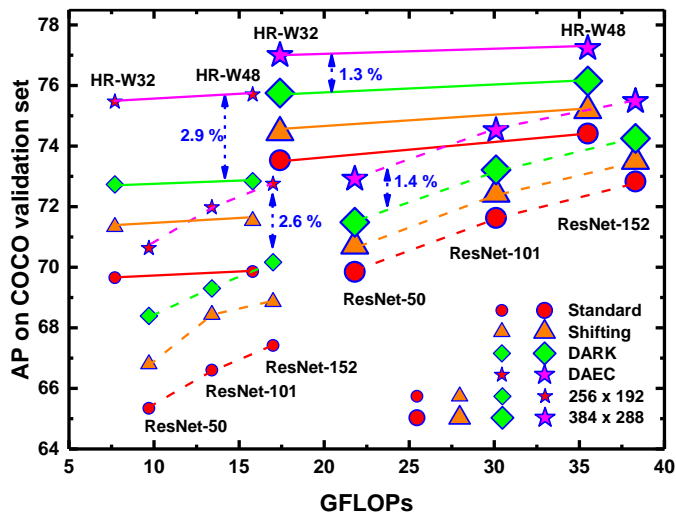

Figure 1. Comparison with the state-of-the-art methods of human pose estimation on the COCO validation dataset. Solid and dash lines denote the HRNet group and the ResNet group, respectively. Tested with a variety of model architectures and input sizes, the proposed method overwhelmingly outperforms other competitors by significant accuracy gains with negligible extra computation.

## 1. Introduction

Being a fundamental computer vision task, human pose estimation (Andriluka et al. 2014) serves as the footstone of more comprehensive human understanding tasks such as intention speculation (Heilbron et al. 2015) and intelligent security (Karanam et al. 2019). Early research on human pose estimation employs graph partition to detect limbs (Pishchulin et al. 2016), which is tedious and relatively ineffective. In recent years, the rapid advance in neural network and deep learning (Rumelhart et al. 1988) has tremendously boosted the performance of both single person (Newell et al. 2016) and multi-person pose estimation (Su et al. 2019).

Different from most state-of-the-art methods (Chen et al. 2018b; Li et al. 2019; Lifshitz et al. 2016; Sun et al. 2019; Zhou et al. 2019), who specialize in better neural network architectures, we pay attention to the post-processing phase which has been proved to be significant but largely ignored (Huang et al. 2019; Zhang et al. 2019). Unlike other computer vision tasks like image classification (Deng et al. 2009; He et al. 2016; Szegedy et al. 2015), object detection (He et al. 2018; Liu et al. 2016; Redmon et al. 2015) and semantic segmentation (Chen et al. 2018a; He et al. 2018; Ronneberger et al. 2015), the pose estimation task employs metrics that are more sensitive to post-processing because they compare ground truth human joint coordinates with model predicted coordinates. Thus, it is of significant importance to study the post-processing phase in depth and develop accurate and reliable methods.

What post-processing does, in pose estimation tasks, is to extract human joint coordinates from neural network outputs, which is also termed as coordinate *decoding* (Zhang et al. 2019). Generally speaking, the widely used neural network based pose estimation methods fall into two categories: coordinate regression (Carreira et al. 2016; Fan et al. 2015; Sun et al. 2018; Toshev et al. 2014) and heatmap regression (Chen et al. 2018b; Li et al. 2019; Sun et al. 2019; Tompson et al. 2014; Xiao et al. 2018). For coordinate regression, coordinate decoding is simple and straightforward because the joint coordinates, or their simple functions, are directly output by the regression model. While suffering from relatively lower accuracy, coordinate regression is employed by only handful algorithms (Carreira et al. 2016; Fan et al. 2015; Sun et al. 2018; Toshev et al. 2014) and is not discussed in current work. By contrast, being the *de facto* standard of human pose estimation, the heatmap regression is more sophisticated and effective owing to preserving context information during forward propagation. That is why it is extensively employed by state-of-the-art methods (Chen et al. 2018b; Li et al. 2019; Sun et al. 2019; Tompson et al. 2014; Xiao et al. 2018) and still remains the optimal option for pose estimation.

Heatmap decoding[1] is to estimate joint coordinates from predicted heatmaps, whose corresponding ground truths subject to a certain distribution (*e.g.* Gaussian distribution). Although directly determining joint coordinate predictions and impacting model performance, unfortunately, the importance of heatmap decoding is generally underestimated (Zhang et al. 2019). In light of that, most state-of-the-art methods suffer from two common problems in their heatmap decoding process: (1) the heatmap information is not sufficiently exploited (Newell et al. 2016); (2) the errors introduced by defective methods are ignored (Zhang et al. 2019).

We propose a novel coordinate decoding method named ***D**istribution-**A**ware and **E**rror-**C**ompensation Coordinate Decoding* (DAEC), which elegantly addresses the above limitations and outperform previous decoding methods (Newell et al. 2016; Zhang et al. 2019) with negligible extra computation. Moreover, the proposed algorithm serves as an efficient plug-in and can be readily integrated with heatmap based methods without changing model designs. Specifically, inspired by the principle that adequate heatmap information leads to accurate prediction (Newell et al. 2016), we evaluate joint coordinates by integrating over a maximal-centered heatmap subarea, which is sufficiently large to provide adequate probability distribution information. Moreover, heatmaps come with noises; we theoretically proved that those errors can be elegantly compensated by tailoring the integral region by an appropriately margin learned from training data. As illustrated in Figure 1 and Figure 8, tested with a variety of models and input sizes, DAEC outperforms its competitors by notable margins, validating the generality and accuracy of the proposed method.

The main contributions of this work can be summarized as follows:

(1) *Distribution-Aware*: By analyzing the previous decoding methods, we found that more accurate method tends to involve more probability distribution information on heatmap. We take a step further and use the entire heatmap for decoding, which, indeed, leads to dramatic accuracy gains.
(2) *Error-Compensation*: Interestingly, the existing decoding methods assuming no errors. We experimentally proved that heatmap based approach brings biased errors and formulated an approach to alleviate those errors by introducing an error-compensation factor, which is learned from training data.
(3) *State-of-the-art performance*: Extensive experiments validate the generality and accuracy of the proposed method. DAEC beats its competitors by significant margins and sets new state-of-the-art on COCO and MPII benchmarks.
(4) *Fast*: By elaborately tailoring computational load, DAEC is not only more accurate but also significantly faster than the previous best decoding algorithm.
(5) *Generality*: Not confined to human pose estimation, DAEC is applicable to any heatmap based tasks.

---

[1] In this work, we use *heatmap decoding* to represent extracting coordinates from heatmaps and *coordinate decoding* to represent extracting coordinates from arbitrary model outputs. For heatmap based methods, they refer to the same process and thus not intentionally distinguished when there is no ambiguity.

## 2. Related Work

### 2.1 Pose Estimation

Benefiting from the dramatic advance in neural network technique (Rumelhart et al. 1988), pose estimation has entered a new era of rapid development. Human pose estimation is commonly split into single and multi-person tasks.

Without learning joint connection knowledge, single person pose estimation detects only human joints and, as a consequence, achieves relatively high performance (Li et al. 2019; Sun et al. 2019; Wei et al. 2016; Xiao et al. 2018). Multi-person pose estimation further falls into two categories: top-down methods (Chen et al. 2018b) and bottom-up methods (Cao et al. 2017; Newell et al. 2017; Sekii 2018).

Top-down methods are essentially integrating person detection with single person pose estimation. Bounding boxes of person instances are first detected by a person detector, such as YOLO (Redmon et al. 2015) and Mask RCNN (He et al. 2018), and then those persons are cropped and fed to a single person pose estimator like CPN (Chen et al. 2018b) and HRNet (Sun et al. 2019). Such two-stage process, to a certain extent, suffers from inference speed, while brings effective performance in return. Most state-of-the-art performance on multi-person pose estimation challenges are achieved with top-down framework (Li et al. 2019).

Bottom-up methods detect multi-person in one shot by learning not only the joints coordinates but also the limb connections (Cao et al. 2017; Newell et al. 2017; Sekii 2018). Representative connection learning designs include learning the part affinity fields (Cao et al. 2017), grouping human joints by associate embedding (Newell et al. 2017) and learning connections by probabilities (Sekii 2018).

### 2.2 Heatmap Decoding

Unlike network design research, heatmap decoding is a largely ignored perspective in the literature. The standard, as well as the most widely used, heatmap decoding method simply extracts the maximal coordinates after smoothing the heatmap with Gaussian filter (Tompson et al. 2014). The standard decoding suffers from two issues:

(1) The heatmap information is extensively wasted by taking only the maximal activation.
(2) The Gaussian smoothing removes only random noises and the more significant biased errors are directly omitted.

To alleviate issue (1) to a certain extent, Newell et al. (Newell et al. 2016) proposed an empirical method that locates joint coordinates by shifting the maximal towards the second maximal by ¼ the distance between them. Involving two instead of one maximals of the heatmap, the shifting method exceeds the standard method marginally. To take a step further, Zhang et al. (Zhang et al. 2019) proposed a decoding method, named DARK, which taking first and second order derivatives on the heatmap to solve the mean value of Gaussian distribution. By further exploiting the heatmap, the DARK method presents better prediction but still suffers from issue (2) mentioned above.

In contrast to all previous works, we tackle the above limitations by proposing a distribution-aware and error-compensation heatmap decoding method. It is proved to be effective theoretically and experimentally. Besides, it is worth noting that the proposed method serves as a lightweight model-agnostic plug-in, which can be seamlessly integrated with existing models without changing network architectures.

## 3. Methodology

In this section, we first briefly review the heatmap encoding process (*i.e.* transforming coordinates into heatmaps) in **Section 3.1** and describe the proposed distribution-aware and error-compensation heatmap decoding method at length in **Section 3.2** before the algorithm implementation is detailed in **Section 3.3**.

### 3.1 Heatmap Encoding

For heatmap based human pose estimation, human joint coordinates are encoded into heatmaps proportionally in terms of their relative locations in the input image.

$$p' = \lambda p \qquad (1)$$

where $p$ and $p'$ denote the point coordinates before and after encoding respectively; $\lambda$ denotes the output stride of neural networks. The most straightforward encoding is to set the very joint pixel as one with others as zero. However, with only one informative pixel, heatmap reproducing is challenging and error-prone. By contrast, learning continuous probability distribution is a relatively easier and more robust process for neural networks. In light of that, encoding human joint coordinates as probability distributions has almost become a *de facto* standard for human pose estimation. Most commonly, a 2D-Gaussian distribution centered at the joint pixel, as shown below.

$$G(p; \mu, \Sigma) = \frac{1}{2\pi |\Sigma|^{\frac{1}{2}}} \exp(-\frac{1}{2}(p - \mu)^T \Sigma^{-1}(p - \mu)) \tag{2}$$

where $\mu$ is the mean value and $\Sigma$ is the covariance matrix.

### 3.2 The Distribution-Aware and Error-Compensation Heatmap Decoding

#### 3.2.1 Distribution-Aware

Errors of human pose estimation are introduced from two sources: *data processing* and *neural network fitting*. On one hand, data processing introduces quantization errors which are to some extent biased. On the other hand, neural networks specialize in function fittings, so they inevitably introduce fitting errors, which can be biased and random. We consider all these errors as noises and aim at accurately determining human joint coordinates from noisy heatmap predictions.

Figure 2 compares maximal-aware and distribution-aware decoding methods. By mapping the pixel with the maximum activation back to the input image, the maximal-aware decoding yields only integer coordinates, causing biased errors toward the upper-left corner. By contrast, considering vicinal probability distributions, the distribution-aware decoding predicts coordinates with decimals, leading to better performance. The shifting (Newell et al. 2016) and DARK (Zhang et al. 2019) methods are only weakly distribution-aware for only partial heatmap information is involved.

In this work, we implement strong distribution-awareness by integrating over the entire heatmap to predict the weighted human joint location (formulated in **section 3.2.2**). While alleviating biased errors to a certain extent, the distribution-aware decoding still cannot thoroughly present biased-error-free predictions before our error-compensation approach (detailed in **section 3.2.2**) takes over.

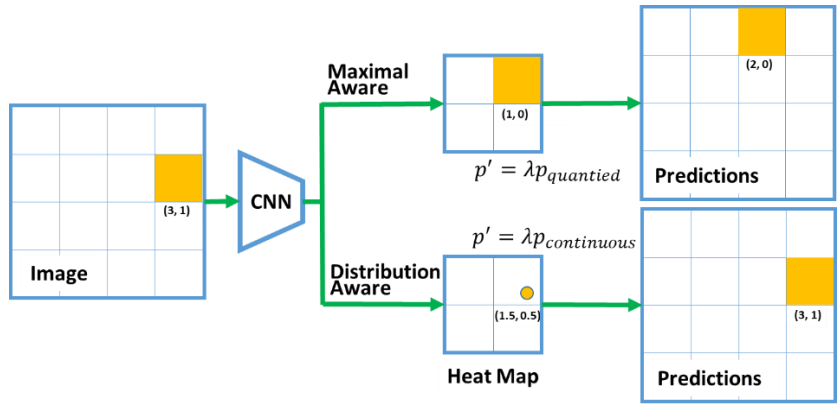

Figure 2. Comparison between maximal-aware and distribution-aware decoding. By mapping the pixel with the maximum activation back to the input image, the maximal-aware decoding yields only integer coordinates, causing biased errors toward the upper-left corner. By contrast, considering vicinal probability distributions, the distribution-aware decoding predicts coordinates with decimals, leading to better performance. While alleviating biased errors to a certain extent, the distribution-aware decoding still cannot thoroughly present biased-error-free predictions before our error-compensation approach takes over.

#### 3.2.2 Error-Compensation

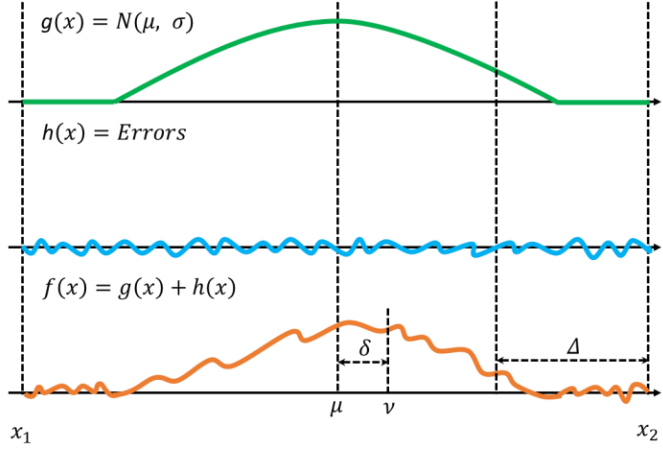

Figure 3. Signal and noise functions of heatmaps. $g(x)$ denotes the signal function, it can be Gaussian distribution or any other distribution used for ground truth heatmap generation; $h(x)$ denotes the noise function containing both random and biased errors; $f(x)$ denotes the predicted heatmap from neural network. The objective of decoding methods is to accurately solve the ground truth mean value $\mu$ from the noisy heatmap $f(x)$.

This section formulates our error-compensation strategy on the basis of distribution-aware integral. Figure 3 anatomizes the predicted heatmap from neural networks. For simplicity, we illustrate distributions in 1D and the conclusions can be readily applied to 2D scenarios. We represent the predicted heatmap as

$$f(x) = g(x) + h(x) \tag{3}$$

where $g(x)$ denotes the signal function, it can be Gaussian distribution or any other distribution used for ground truth heatmap generation; $h(x)$ denotes the noise function containing both random and biased errors and $f(x)$ denotes the predicted heatmap from neural network. The objective of decoding methods is to accurately solve ground truth mean value $\mu$ from the noisy heatmap $f(x)$. According to the definition of mean value, $\mu$ can be calculated using

$$\mu = \int_{x_1}^{x_2} xg(x)dx \Big/ \int_{x_1}^{x_2} g(x)dx \tag{4}$$

Similarly the mean value $v$ of $f(x)$ can be obtained by

$$v = \int_{x_1}^{x_2} xf(x)dx \Big/ \int_{x_1}^{x_2} f(x)dx \tag{5}$$

By substituting Eq. (3) into Eq. (5), $v$ can be expressed as

$$v = \int_{x_1}^{x_2} x[g(x) + h(x)]dx \Big/ \int_{x_1}^{x_2} g(x) + h(x)dx$$

$$= \left\{\int_{x_1}^{x_2} xg(x)dx + \int_{x_1}^{x_2} xh(x)dx\right\} \Big/ \left\{\int_{x_1}^{x_2} g(x)dx + \int_{x_1}^{x_2} h(x)dx\right\} \tag{6}$$

for a well converged neural network, the signal-noise ratio should be large enough, thus the integral of noise is negligible compared with that of the signal. Therefore

$$\int_{x_1}^{x_2} g(x)dx \gg \int_{x_1}^{x_2} h(x)dx \tag{7}$$

By substituting Eq. (7) into Eq. (6), $v$ can be approximated as

$$v \approx \left\{\int_{x_1}^{x_2} xg(x)dx + \int_{x_1}^{x_2} xh(x)dx\right\} \Big/ \int_{x_1}^{x_2} g(x)dx$$

$$= \mu + \int_{x_1}^{x_2} xh(x)dx \Big/ \int_{x_1}^{x_2} g(x)dx$$

$$= \mu + \delta \tag{8}$$

where

$$\delta = \int_{x_1}^{x_2} xh(x)dx \Big/ \int_{x_1}^{x_2} g(x)dx \tag{9}$$

Even evaluating $\delta$ is difficult, according to the continuity of integral, we can always find a value $\Delta$ which satisfies the formula below

$$\mu = \nu - \delta = \int_{x_1}^{x_2-\Delta} xf(x)dx \Big/ \int_{x_1}^{x_2-\Delta} f(x)dx \tag{10}$$

Therefore, given the predicted heatmap $f(x)$, the optimal estimation of joint coordinate $\mu$ can be obtained by sliding $\Delta$ around $x_2$. For instance, if $\delta > 0$, which means $\nu$ is on the right-hand side of $\mu$. By reducing the upper bound of integral region from $x_2$ to $x_2 - \Delta$, $\nu$ declines gradually, and eventually meets $\nu = \mu$ at a specific value of $\Delta$. In this work, we refer $\Delta$ as the *error-compensation factor*, which is learned from the training dataset.

After a model is well-converged on training dataset, we evaluate the model performance on the training dataset with a range of $\Delta$ values (*e.g.* $\Delta \in \{..., -2, -1, 0, 1, 2 ...\}$). The one achieves the best performance is the optimal error-compensation factor (denoted as $\Delta_{opt}$), which will be used in test scenarios.

While formulated with 1D, the above derivation works for 2D scenarios as well. Besides, treated equally during data encoding, network inference and coordinate decoding, the $x$ and $y$ dimensions subject to qualitatively identical error distribution, suggesting that the error-compensation approach can be approximately simplified by using the same $\Delta$ for both axes.

*Remarks:* DAEC features four advantages. (1) Involving sufficient probability distribution information, DAEC fully exploits heatmaps, assisting in locating more reasonable sub-pixel coordinates and reducing quantization errors. (2) By tailoring the integral region, DAEC compensates the overall errors in one-shot regardless of error sources and patterns. (3) Unlike previous Gaussian-smooth-sensitive methods, DAEC is smoothing-free because integral operation is suitable for any surfaces even with noises and both biased and random errors are indiscriminately compensated. **Section 4.5** discusses the effect of smoothing in depth.

### 3.3 Implementation Details

The optimal compensation factor $\Delta_{opt}$ is learned by evaluating model performance on the training dataset with a range of $\Delta_{opt}$ candidates (*e.g.* $\Delta \in \{..., -2, -1, 0, 1, 2 ...\}$), following the strategy depicted in Figure 4 and Figure 5, specifically, by four steps.

(1) Locate the maximal (point A in Figure 4) of the heatmap.
(2) Expand the maximal-centered integral region to $6\sigma + 3$, where $\sigma$ is the standard deviation of Gaussian distribution. An approximately Gaussian distribution obeys the $3\sigma$-region rule that the values within three standard deviations ($[-3\sigma, 3\sigma]$) account for about 99.7% of the set. To further cover the entire distribution, the borders are expanded by an extra pixel at each end reaching $6\sigma + 3$. Note that a larger region also works but brings extra computation.
(3) Reduce the upper bound by $\Delta$ for both axes (only show one axis in Figure 4) and integrate over the region with Eq. (10) to get the predicted joint coordinates and evaluate the model performance (*e.g.* AP on COCO) with these predictions.
(4) Lastly, the $\Delta$ achieving the best model performance is the optimal error-compensation factor $\Delta_{opt}$ which will be used in test scenarios.

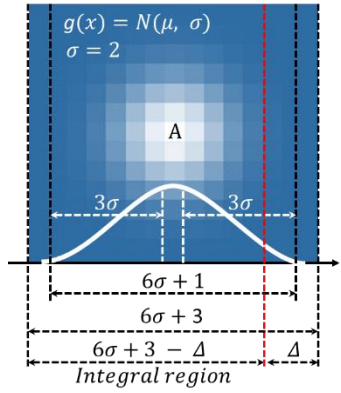

Figure 4. The Strategy of generating integral region and learning optimal compensation factor $\Delta$ from predicted heatmaps of neural networks. $\sigma$ is the standard deviation of Gaussian distribution. An approximately Gaussian distribution obeys the $3\sigma$-region rule that the values within three standard deviations ($[-3\sigma, 3\sigma]$) account for about 99.7% of the set. To further cover the entire distribution, the borders are expanded by an extra pixel at each end reaching $6\sigma + 3$. Note that a larger region also works but brings extra computation.

$$
\begin{aligned}
&Candidates = [\ldots, -2, -1, 0, 1, 2 \ldots] \\
&Accuracies = \{\,\} \\
&for\ \Delta\ in\ Candidates: \\
&\quad Accuracy = evaluate(TrainingSet,\ \Delta) \\
&\quad Accuracies[\Delta] = Accuracy \\
&\Delta_{opt} = \underset{\Delta}{argmax}(Accuracies)
\end{aligned}
$$

Figure 5. Pseudo code to determine $\Delta_{opt}$ from training dataset. We evaluate a set of $\Delta_{opt}$ candidates on training dataset and pick the one leads to the best model performance as $\Delta_{opt}$, which will be used in test scenarios.

Figure 6 illustrates the effect of the error-compensation factor $\Delta$ on various models on the COCO training set. The $\Delta_{opt}$ value is the one leads to the best model performance, particularly in Figure 6, $\Delta_{opt} = 4$ for $\sigma = 2$ and $\Delta_{opt} = 5$ for $\sigma = 3$. We further conclude an empirical formula of $\Delta_{opt} = \sigma + 2$ for these cases, suggesting that the biased errors are approximately determined by the size of Gaussian filter, *i.e.* larger filter introduces larger biased error and, as a result, a larger $\Delta$ is required for error compensation.

Note that we use the term "learn" for $\Delta_{opt}$ determination because $\Delta_{opt}$ is task-specific, which means this hyper-parameter $\Delta_{opt}$ is "learned" from a specific dataset for specific a model. Even although, an empirical formula can be concluded, its generality requires further validations. We instead recommend the task-specific $\Delta_{opt}$ value learned for each individual case.

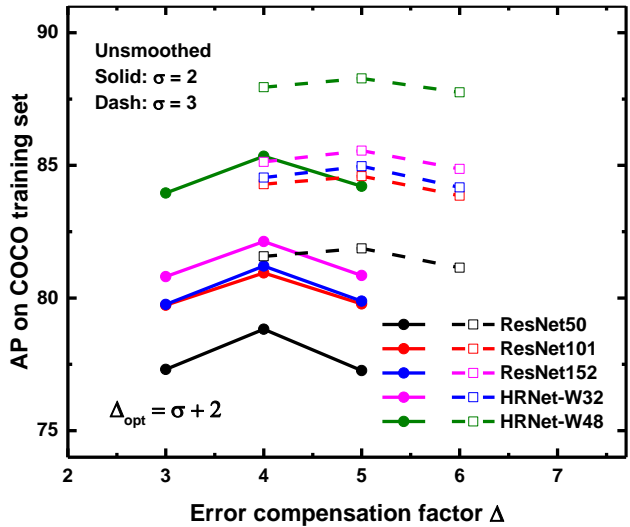

Figure 6. Effect of the error-compensation factor $\Delta$ on the state-of-the-art methods of human pose estimation on the COCO training set, evaluated with unsmoothed heatmaps. As can be conclude, regardless of neural networks architectures, the maximal is always at $\Delta_{opt} = \sigma + 2$.

# 4 Experiments

## 4.1 Experiment Settings

**Datasets and Metrics:** Two widely used human pose estimation datasets: COCO-2017 and MPII are employed for extensive tests. The COCO keypoint dataset (Lin et al. 2014) contains 200,000 images of more than 250,000 person samples with various body scales, background environments and occlusion patterns. Each person instance is labelled with 17 joints. The MPII human pose dataset (Andriluka et al. 2014) contains 20,000 images with more than 40,000 person samples, each labeled with 16 keypoints. Object Keypoint Similarity (OKS) and Percentage of Correct Points (PCK) metrics are respectively used for the COCO and MPII datasets to evaluate model performance.

**Neural Networks:** The High-Resolution network groups (HR-W32 and HR-W48) (Sun et al. 2019) and Simple-Baseline network groups (ResNet-50, ResNet-101 and ResNet-152) (Xiao et al. 2018) are tested with three different input sizes (128 × 96, 256 × 192, 384 × 288). We follow the same data processing and training strategies as the original papers.

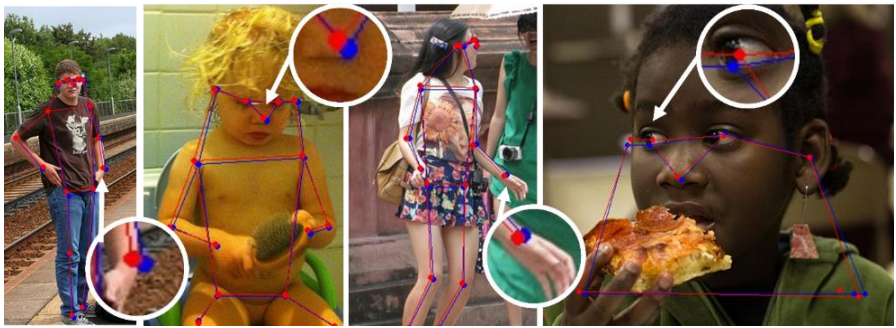

Figure 7. Qualitative evaluation of the HR-W32-256×192 model with (red) and without (blue) DAEC on COCO. When equipped with DAEC, the model predicts more reasonable joint coordinates.

## 4.2 Results on COCO

Figure 1 compares the proposed approach with the state-of-the-art methods of human pose estimation on the COCO validation dataset. Tested with a variety of model architectures and input sizes, DAEC outperforms previous methods by considerable accuracy gains. Specifically, as listed in Table 1, the proposed approach overwhelmingly exceeds its competitors under a variety of metrics.

Take the most representative AP (average precision) metric as an example, the Simple-Baseline ResNet-50, ResNet-101 and ResNet-152 models remarkably gain 2.24, 2.68, and 2.59 AP, respectively, at the size of 256×192. Even more surprisingly, the accuracies of HR-W32 and HR-W48 are increase by 2.73 and 2.86 AP reaching 75.47 and 75.70 AP, respectively. Averagely, the AP accuracy is promoted by 2.62 AP over the previous state-of-the-art method (Zhang et al. 2019). Similarly, slightly lower than the 256×192 input scenario, but still significant improvements can be observed at the input size of 384×288. Qualitative evaluation, shown in Figure 7, demonstrates that equipped with DAEC, the model tends to predict more reasonable joint coordinates.

Table 1. Comparison with the state-of-the-art methods of human pose estimation on the COCO validation dataset[†]. Validated with a variety of models and input sizes, DAEC overwhelmingly outperforms previous methods by considerable accuracy gains under a variety of metrics.

| Model | Input | Method | AP | AP[†‡] | AP$^{50}$ | AP$^{75}$ | AP$^M$ | AP$^L$ | AR | AR$^{50}$ | AR$^{75}$ | AR$^M$ | AR$^L$ |
|---|---|---|---|---|---|---|---|---|---|---|---|---|---|
| ResNet-50 | 256×192 | Standard | 65.34 | 5.29↑ | 90.37 | 74.48 | 63.25 | 68.59 | 69.32 | 91.85 | 77.96 | 66.57 | 73.48 |
| ResNet-50 | 256×192 | Shifting | 66.80 | 3.83↑ | 90.43 | 75.74 | 65.15 | 70.28 | 70.84 | 91.99 | 78.90 | 68.09 | 75.00 |
| ResNet-50 | 256×192 | DARK | 68.40 | 2.24↑ | 91.38 | 76.89 | 66.60 | 71.59 | 72.01 | 92.07 | 79.72 | 69.30 | 76.14 |
| ResNet-50 | 256×192 | **DAEC** | **70.63** |  | **91.40** | **78.17** | **68.27** | **74.66** | **74.11** | **92.24** | **80.81** | **70.98** | **78.85** |
| ResNet-50 | 384×288 | Standard | 69.85 | 3.07↑ | 91.46 | 77.07 | 66.86 | 74.66 | 73.28 | 92.48 | 79.83 | 69.55 | 78.80 |
| ResNet-50 | 384×288 | Shifting | 70.71 | 2.21↑ | 91.47 | 78.01 | 67.45 | 75.55 | 73.96 | 92.51 | 80.26 | 70.18 | 79.56 |
| ResNet-50 | 384×288 | DARK | 71.49 | 1.43↑ | 91.47 | 78.20 | 68.43 | 76.50 | 74.71 | 92.66 | 80.79 | 70.93 | 80.35 |
| ResNet-50 | 384×288 | **DAEC** | **72.92** |  | **91.52** | **79.41** | **69.20** | **78.45** | **75.80** | **92.87** | **81.72** | **71.72** | **81.86** |
| ResNet-101 | 256×192 | Standard | 66.60 | 5.38↑ | 91.45 | 75.77 | 65.21 | 69.60 | 70.54 | 92.46 | 78.84 | 68.04 | 74.35 |

| Backbone | Input size | Method | AP | AP↑ | | | | | | | | | |
|---|---|---|---|---|---|---|---|---|---|---|---|---|---|
| ResNet-101 | 256×192 | Shifting | 68.43 | 3.55↑ | 91.44 | 77.89 | 66.77 | 71.40 | 72.06 | 92.44 | 80.05 | 69.60 | 75.86 |
| ResNet-101 | 256×192 | DARK | 69.30 | 2.68↑ | 91.48 | 78.08 | 67.85 | 72.60 | 73.13 | 92.66 | 80.72 | 70.66 | 76.99 |
| ResNet-101 | 256×192 | **DAEC** | **71.98** | | **92.48** | **79.32** | **69.60** | **75.73** | **75.31** | **93.15** | **81.85** | **72.44** | **79.73** |
| ResNet-101 | 384×288 | Standard | 71.63 | 2.89↑ | 92.44 | 80.19 | 69.04 | 76.02 | 75.07 | 93.25 | 82.24 | 71.75 | 80.12 |
| ResNet-101 | 384×288 | Shifting | 72.42 | 2.10↑ | 92.45 | 80.25 | 69.78 | 76.66 | 75.76 | 93.26 | 82.51 | 72.49 | 80.75 |
| ResNet-101 | 384×288 | DARK | 73.22 | 1.31↑ | 92.47 | 80.35 | 70.70 | 77.68 | 76.51 | 93.31 | 82.97 | 73.20 | 81.56 |
| ResNet-101 | 384×288 | **DAEC** | **74.52** | | **92.47** | **81.40** | **71.44** | **79.40** | **77.55** | **93.42** | **83.61** | **73.97** | **82.99** |
| ResNet-152 | 256×192 | Standard | 67.42 | 5.34↑ | 91.48 | 76.75 | 65.51 | 70.85 | 71.26 | 92.66 | 79.83 | 68.63 | 75.28 |
| ResNet-152 | 256×192 | Shifting | 68.86 | 3.90↑ | 91.52 | 77.86 | 67.10 | 72.23 | 72.60 | 92.85 | 80.68 | 70.02 | 76.55 |
| ResNet-152 | 256×192 | DARK | 70.17 | 2.59↑ | 92.47 | 78.93 | 68.17 | 73.59 | 73.74 | 93.03 | 81.27 | 71.13 | 77.77 |
| ResNet-152 | 256×192 | **DAEC** | **72.75** | | **92.51** | **80.34** | **70.00** | **76.84** | **75.95** | **93.14** | **82.68** | **72.84** | **80.68** |
| ResNet-152 | 384×288 | Standard | 72.83 | 2.65↑ | 92.50 | 81.38 | 70.24 | 76.99 | 76.15 | 93.64 | 83.50 | 72.95 | 81.00 |
| ResNet-152 | 384×288 | Shifting | 73.51 | 1.98↑ | 92.52 | 81.47 | 70.96 | 77.74 | 76.80 | 93.73 | 83.80 | 73.60 | 81.67 |
| ResNet-152 | 384×288 | DARK | 74.26 | 1.23↑ | **92.54** | 82.44 | 71.88 | 78.63 | 77.50 | 93.77 | 84.32 | 74.34 | 82.31 |
| ResNet-152 | 384×288 | **DAEC** | **75.48** | | **92.54** | **82.59** | **72.57** | **80.33** | **78.50** | **93.84** | **84.70** | **75.05** | **83.75** |
| HR-W32 | 256×192 | Standard | 69.66 | 5.81↑ | 92.49 | 79.02 | 67.87 | 73.16 | 73.42 | 93.77 | 81.99 | 70.79 | 77.48 |
| HR-W32 | 256×192 | Shifting | 71.33 | 4.13↑ | 92.49 | 81.11 | 69.63 | 74.68 | 74.85 | 93.78 | 83.01 | 72.21 | 78.95 |
| HR-W32 | 256×192 | DARK | 72.74 | 2.73↑ | 92.51 | 81.41 | 70.85 | 76.57 | 76.24 | 93.83 | 83.82 | 73.46 | 80.53 |
| HR-W32 | 256×192 | **DAEC** | **75.47** | | **93.49** | **83.50** | **72.86** | **79.52** | **78.35** | **94.05** | **85.11** | **75.26** | **83.13** |
| HR-W32 | 384×288 | Standard | 73.53 | 3.47↑ | 92.54 | 82.21 | 71.24 | 77.74 | 76.94 | 93.88 | 84.15 | 73.69 | 81.92 |
| HR-W32 | 384×288 | Shifting | 74.45 | 2.55↑ | 92.54 | 82.33 | 71.84 | 78.62 | 77.69 | 93.92 | 84.49 | 74.45 | 82.66 |
| HR-W32 | 384×288 | DARK | 75.75 | 1.25↑ | **93.55** | 83.33 | 73.05 | 79.92 | 78.71 | **94.16** | 85.06 | 75.45 | 83.72 |
| HR-W32 | 384×288 | **DAEC** | **77.00** | | 93.54 | **83.67** | **73.86** | **81.86** | **79.71** | 94.14 | **85.64** | **76.17** | **85.13** |
| HR-W48 | 256×192 | Standard | 69.86 | 5.85↑ | 92.48 | 79.79 | 68.12 | 73.31 | 73.70 | 93.73 | 82.31 | 70.90 | 77.92 |
| HR-W48 | 256×192 | Shifting | 71.53 | 4.17↑ | 92.50 | 81.03 | 69.56 | 75.05 | 75.23 | 93.78 | 83.28 | 72.38 | 79.55 |
| HR-W48 | 256×192 | DARK | 72.84 | 2.86↑ | 92.52 | 82.11 | 71.18 | 76.36 | 76.51 | 93.86 | 84.18 | 73.70 | 80.81 |
| HR-W48 | 256×192 | **DAEC** | **75.70** | | **93.50** | **83.56** | **73.05** | **79.92** | **78.71** | **94.07** | **85.53** | **75.44** | **83.68** |
| HR-W48 | 384×288 | Standard | 74.42 | 2.82↑ | 93.48 | 82.41 | 71.72 | 78.60 | 77.60 | 94.05 | 84.65 | 74.41 | 82.49 |
| HR-W48 | 384×288 | Shifting | 75.18 | 2.05↑ | 93.48 | 82.53 | 72.54 | 79.39 | 78.28 | 94.11 | 84.93 | 75.11 | 83.16 |
| HR-W48 | 384×288 | DARK | 76.15 | 1.08↑ | 93.50 | 83.69 | 73.59 | 80.46 | 79.15 | 94.11 | 85.67 | 75.99 | 84.02 |
| HR-W48 | 384×288 | **DAEC** | **77.23** | | **93.52** | **83.74** | **74.15** | **82.25** | **80.07** | **94.24** | **85.97** | **76.61** | **85.41** |

† Evaluated without flip.

‡ AP↑ stands for DAEC accuracy gain over other methods.

### 4.3 Results on MPII

Figure 8 compares the proposed method with the state-of-the-art methods of human pose estimation on the MPII dataset. Still, the proposed method outperforms previous methods by considerable accuracy gains for various network architectures and input sizes. Particularly, for the more rigorous PCKh$^{0.1}$ metric, DAEC incredibly promotes the previous state-of-the-art performance by 8.4% and 7.8% for HR-W32-256×256 and ResNet-152-256×256 frameworks, respectively (subgraph on the left). When evaluated by the less strict PCHKh$^{0.5}$ metric, the improvement gap shrinks due to loose and less distinguishable standard (subgraph on the right). Table 2 lists the average and joint-specific PCKh values, as can be noted, the state-of-the-art model performance are significantly improved by integrating with our method.

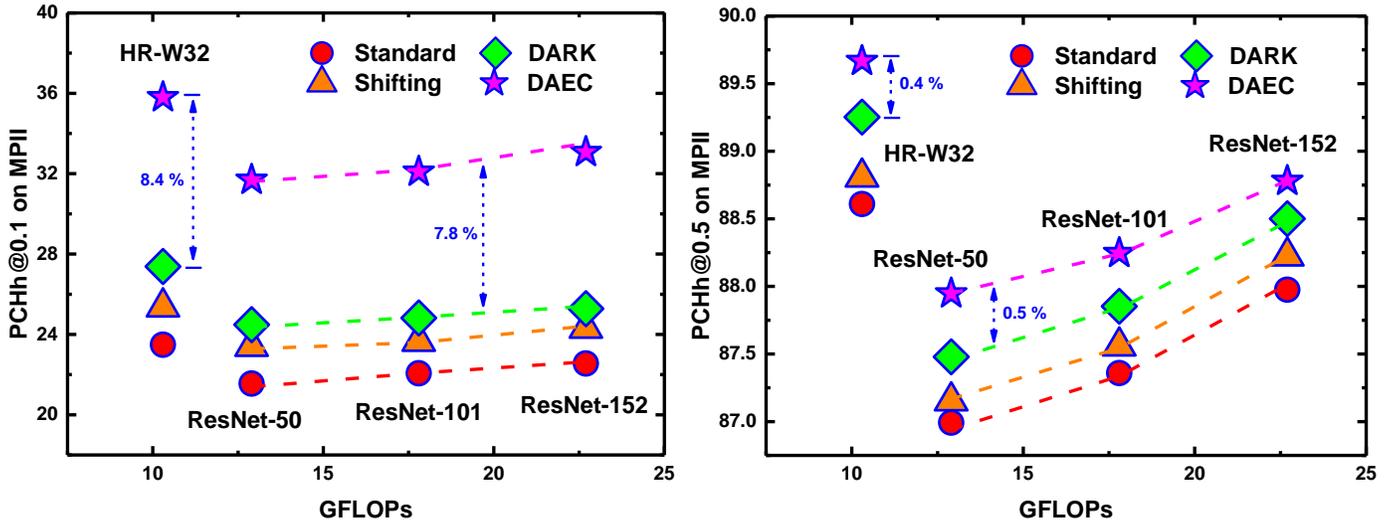

Figure 8. Comparison with the state-of-the-art methods of human pose estimation on the MPII dataset. Tested with a variety of model architectures and input sizes, the proposed method outperforms previous methods by considerable accuracy gains.

Table 2. Comparison with the state-of-the-art methods of human pose estimation on the MPII dataset[†]. The proposed method outperforms previous methods under a variety of metrics, model architectures and input sizes.

| Model | Method | Head | Shoul. | Elbow | Wrist | Hip | Knee | Ankle | PCKh$^{0.1}$ | ↑[‡] | PCKh$^{0.5}$ | ↑[‡] |
|---|---|---|---|---|---|---|---|---|---|---|---|---|
| ResNet-50 | Standard | 96.04 | 94.19 | 87.25 | 81.34 | 86.15 | 81.60 | 78.32 | 21.55 | 10.13↑ | 86.99 | 0.95↑ |
| ResNet-50 | Shifting | 96.04 | 94.34 | 87.35 | 81.53 | 86.41 | 81.85 | 78.48 | 23.40 | 8.28↑ | 87.15 | 0.79↑ |
| ResNet-50 | DARK | **96.15** | 94.53 | 87.76 | 81.87 | 86.76 | 82.49 | 78.81 | 24.48 | 7.20↑ | 87.48 | 0.47↑ |
| ResNet-50 | **DAEC** | 95.87 | **94.87** | **88.44** | **82.05** | **87.62** | **83.22** | **79.48** | **31.69** | | **87.95** | |
| ResNet-101 | Standard | 96.35 | 94.62 | 87.40 | 82.41 | 85.72 | 82.35 | 78.77 | 22.07 | 10.02↑ | 87.36 | 0.89↑ |
| ResNet-101 | Shifting | **96.59** | 94.58 | 87.69 | 82.39 | 86.22 | 82.71 | 78.98 | 23.66 | 8.43↑ | 87.56 | 0.69↑ |
| ResNet-101 | DARK | 96.32 | 94.72 | 88.07 | 82.85 | 86.71 | 83.16 | 79.24 | 24.82 | 7.27↑ | 87.85 | 0.39↑ |
| ResNet-101 | **DAEC** | 96.28 | **94.80** | **88.55** | **83.42** | **87.54** | **83.42** | **79.74** | **32.09** | | **88.25** | |
| ResNet-152 | Standard | 96.62 | 95.02 | 88.27 | 82.70 | 86.38 | 83.30 | 79.85 | 22.55 | 10.52↑ | 87.98 | 0.80↑ |
| ResNet-152 | Shifting | 96.62 | 95.31 | 88.56 | 82.99 | 86.91 | 83.58 | 79.83 | 24.31 | 8.76↑ | 88.23 | 0.56↑ |
| ResNet-152 | DARK | **96.73** | 95.33 | 88.80 | 83.66 | 87.02 | 83.78 | **80.63** | 25.28 | 7.79↑ | 88.50 | 0.28↑ |
| ResNet-152 | **DAEC** | 96.56 | **95.67** | **88.97** | **83.85** | **87.99** | **84.14** | 80.52 | **33.07** | | **88.78** | |
| HR-W32 | Standard | 96.79 | 95.06 | 89.08 | 84.29 | 86.01 | 84.40 | 81.39 | 23.49 | 12.31↑ | 88.61 | 1.06↑ |
| HR-W32 | Shifting | 96.93 | 95.25 | 89.06 | 84.39 | 86.43 | 84.89 | 81.58 | 25.36 | 10.44↑ | 88.81 | 0.86↑ |
| HR-W32 | DARK | **96.97** | 95.40 | 89.57 | 85.03 | 87.04 | 85.67 | 82.03 | 27.38 | 8.42↑ | 89.25 | 0.41↑ |
| HR-W32 | **DAEC** | 96.86 | **95.58** | **89.98** | **85.49** | **87.83** | **86.18** | **82.59** | **35.80** | | **89.67** | |

[†] The metric used for specific joints is PCKh$^{0.5}$ evaluated without flip;

[‡] ↑ stands for DAEC accuracy gain over other methods.

### 4.4 Speed comparison

Speed matters especially for embedded systems. The Shifting, DARK and DAEC methods are implemented on the basis of the standard decoding method which provides the coordinate of the maximal. Table 3 lists the extra elapsed time compared with the standard method. As can be seen, the shifting method provides the fastest speed, however, at the price of low accuracy. Not only is DAEC more accurate, but also significant faster than the DARK method, which attributes to the well-tailored integral region foresaid in **Section 3.3**.

Table 3. Extra elapsed time compared with the standard method, measured with the HR-W32-256×192 model using Intel Core i7-9700F CPU.

| Method | Shifting | DARK | DAEC |
|---|---|---|---|
| Elapsed time (ms/image) | 0.31 | 3.00 | 1.44 |

## 4.5 Effect of Gaussian Smoothing

Previous methods are sensitive to Gaussian smoothing, as shown in Table 4, without Gaussian smoothing, the model performance of DARK degrades dramatically. By contrast, tackling overall errors in one-shot, DAEC is smoothing-free.

However, For comparison, we still tested DAEC with Gaussian smoothing. As illustrated in Figure 9, trained on the COCO training set, the optimal error-compensation factor can be empirically expressed as $\Delta_{opt} = \sigma + 1$. Compared with the unsmoothed scenario, the optimal error-compensation factor $\Delta_{opt}$ declines by one pixel because a portion of errors are removed by Gaussian smoothing and thus a smaller $\Delta$ is required to compensate the rest of errors.

Table 5 evaluates the effect of smoothing on the proposed method on both COCO dataset. As can be noted, comparable performance is obtained with and without smoothing, suggesting that DAEC functions as a good alternative of Gaussian smoothing. However, unsmoothed DAEC brings computational benefits from two perspectives: no smoothing expense and cheaper integral cost due to larger $\Delta_{opt}$.

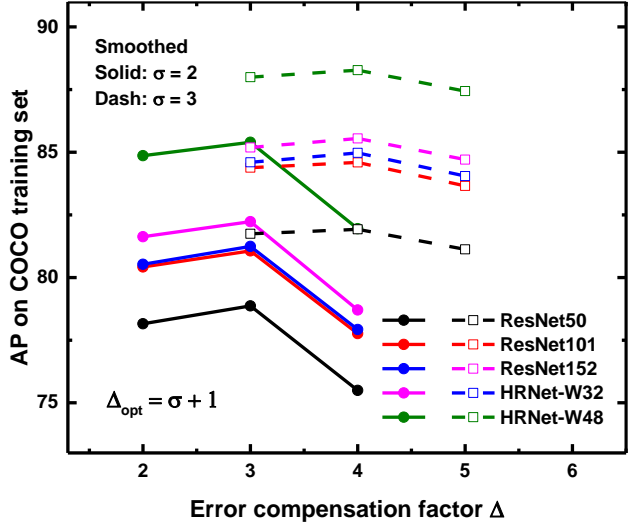

Figure 9. Effect of error-compensation factor $\Delta$ on the state-of-the-art methods of human pose estimation on the COCO training set, evaluated with Gaussian-smoothed heatmaps. Compared with the unsmoothed scenario, the optimal error-compensation factor $\Delta_{opt}$ reduced to $\Delta_{opt} = \sigma + 1$ because a portion of errors are removed by Gaussian smoothing and a smaller $\Delta$ is required for error compensation.

Table 4. Effect of Gaussian smooth on DARK model performances (the AP metric) on the COCO dataset with different network architectures and input sizes.

| Input | Smooth | ResNet-50 | | ResNet-101 | | ResNet-152 | | HR-W32 | | HR-W48 | |
|---|---|---|---|---|---|---|---|---|---|---|---|
| 256×192 | Yes | **68.40** | 0.52↓ | **69.30** | 0.09↓ | **70.17** | 0.26↓ | **72.74** | 0.04↓ | **72.84** | 0.37↓ |
| | No | 67.88 | | 69.21 | | 69.91 | | 72.24 | | 72.47 | |
| 384×288 | Yes | **71.49** | 0.41↓ | **73.22** | 0.24↓ | **74.26** | 0.12↓ | **75.75** | 0.01↓ | **76.15** | 0.43↓ |
| | No | 71.08 | | 72.98 | | 74.13 | | 75.00 | | 75.72 | |

Table 5. Effect of Gaussian smooth on DAEC model performances (the AP metric) on the COCO dataset with different network architectures and input sizes.

| Input | Smooth | ResNet-50 | | ResNet-101 | | ResNet-152 | | HR-W32 | | HR-W48 | |
|---|---|---|---|---|---|---|---|---|---|---|---|
| 256×192 | Yes | **70.69** | 0.06↓ | 71.96 | 0.02↑ | **72.76** | 0.01↓ | 75.43 | 0.04↑ | **75.71** | 0.01↓ |
| | No | 70.63 | | **71.98** | | 72.75 | | **75.47** | | 75.70 | |
| 384×288 | Yes | 72.84 | 0.08↑ | **74.53** | 0.01↓ | 75.31 | 0.17↑ | **77.01** | 0.01↓ | 77.18 | 1.48↑ |
| | No | **72.92** | | 74.52 | | **75.48** | | 77.00 | | **77.23** | |

### 4.6 Effect of Different Error-Compensation Patterns

As analyzed above, the optimal compensation factor $\Delta_{opt}$ is always a positive value, suggesting that the bottom-right (BR) corner of the integral region is cut to compensate errors. We also tested other error-compensation patterns, particularly, upper-left (UL), bottom-left (BR) and upper-right (UR) cuttings. Table 6 and Table 7 compare these four error-compensation patterns on COCO and MPII dataset, respectively. As can be concluded, the model accuracy subjects to BR > UR > BL > UL, suggesting that the model errors universally bias to the bottom-right corner.

Table 6. Comparison of the model performance (AP) of different error-compensation patterns on COCO dataset with different input sizes and model architectures. BR, UL, BL, UR stand for compensating error by cutting the bottom-right, upper-left, bottom-left and upper-right corner of integral region, respectively. As can be concluded, the model accuracy subjects to BR > UR > BL > UL, suggesting that the model errors universally bias to the bottom-right corner.

| Input size | Pattern | ResNet-50 | ResNet-101 | ResNet-152 | HR-W32 | HR-W48 |
|---|---|---|---|---|---|---|
| 256×192 | BR | **70.63** | **71.98** | **72.75** | **75.47** | **75.70** |
|  | UR | 69.05 | 70.39 | 71.21 | 73.90 | 74.14 |
|  | BL | 67.63 | 68.96 | 69.79 | 72.27 | 72.46 |
|  | UL | 66.28 | 67.51 | 68.16 | 70.96 | 71.00 |
| 384×288 | BR | **72.92** | **74.52** | **75.48** | **77.00** | **77.23** |
|  | UR | 71.72 | 73.52 | 74.54 | 76.11 | 76.43 |
|  | BL | 70.93 | 72.64 | 73.76 | 75.10 | 75.61 |
|  | UL | 70.09 | 71.60 | 72.72 | 74.33 | 74.63 |

Table 7. Comparison of the model performance ($PCKh^{0.5}$) of different error-compensation patterns on MPII dataset with different model architectures. BR, UL, BL, UR stand for compensating error by cutting the bottom-right, upper-left, bottom-left and upper-right corner of integral region, respectively. As can be concluded, the model accuracy subjects to BR > UR > BL > UL, suggesting that the model errors universally bias to the bottom-right corner.

| Pattern | ResNet-50 | ResNet-101 | ResNet-152 | HR-W32 |
|---|---|---|---|---|
| BR | **87.95** | **88.25** | **88.78** | **89.67** |
| UR | 87.56 | 87.87 | 88.46 | 89.40 |
| BL | 87.36 | 87.59 | 88.43 | 89.27 |
| UL | 87.05 | 87.36 | 88.07 | 89.00 |

# 5 Conclusions and Future Work

In this work, we studied the largely ignore heatmap decoding process of human pose estimation, finding that heatmap based models intrinsically suffer from biased errors. A novel distribution-aware and error-compensation coordinate decoding method is proposed, which learns its decoding strategy from training data and serves as an effective plug-in with negligible extra computational cost. Compared with previous state-of-the-art methods, the proposed model enables significant accuracy gains regardless of network architectures, input sizes and datasets, suggesting reasonable generality of the current approach. Analysis shows that the errors of heatmap based model universally bias towards the bottom-right corner. Future wrok can be concentrated on validating DAEC with other heatmap based tasks and investigating the generation mechanism of the biased errors.

# 6 Acknowledgements

We thank the reviewers for carefully reading the manuscript and providing very helpful comments. This work was supported in part by the Key-Area Research and Development Program of Guangdong Province (2019B010149002) and Dongguan City Core Technology Research Frontier Project (2019622101001).